\documentclass[lettersize,journal]{IEEEtran}
\usepackage{amsmath,amsfonts}
\usepackage{algorithm}
\usepackage{subcaption}
\usepackage{geometry}
\usepackage{diagbox}
\usepackage{adjustbox}
\usepackage{booktabs}
\usepackage[table]{xcolor}
\usepackage{pgfplots}
\pgfplotsset{width=10cm,compat=1.18}
\usepackage{array}
\usepackage[caption=false,font=normalsize,labelfont=sf,textfont=sf]{subfig}
\usepackage{textcomp}
\usepackage{stfloats}
\usepackage{url}
\usepackage{verbatim}
\usepackage{graphicx}
\usepackage{cite}
\usepackage{tikz}
\usepackage{pgfplots}
\usepackage{multirow, makecell}
\usepackage{algpseudocode}
\usepackage{soul}
\usepackage{xcolor}
\usepackage{listings}
\usetikzlibrary{arrows.meta, positioning, shapes.geometric, backgrounds, fit, calc}
\usetikzlibrary{shapes.geometric, arrows.meta, positioning, shadows.blur, calc}
\usepackage{ragged2e}
\usepackage[capitalize]{cleveref}
\crefname{section}{Sec.}{Secs.}
\Crefname{section}{Section}{Sections}
\Crefname{table}{Table}{Tables}
\crefname{table}{Tab.}{Tabs.}
\pgfplotsset{compat=1.17}

\raggedright
\begin{document}
\justifying % Ensures full justification

\title{A Classification-Aware Super-Resolution Framework for Ship Targets in SAR Imagery}

\author{\IEEEauthorblockN{Ch Muhammad Awais\IEEEauthorrefmark{1}\IEEEauthorrefmark{2}\IEEEauthorrefmark{3},
Marco Reggiannini\IEEEauthorrefmark{2}\IEEEauthorrefmark{3},
Davide Moroni\IEEEauthorrefmark{2}, 
 and
Oktay Karakus\IEEEauthorrefmark{4}}\\
\IEEEauthorblockA{\IEEEauthorrefmark{1}PhD School in Computer Science,
University of Pisa, 56126 Pisa, Italy\\}
\IEEEauthorblockA{\IEEEauthorrefmark{2}Institute of Information Science and Technologies, National Research Council of Italy, 56124 Pisa, Italy,~\IEEEauthorrefmark{3}National Biodiversity Future Center - NBFC, Palermo, Italy,~\IEEEauthorrefmark{4}School of Computer Science and Informatics, Cardiff University, Cardiff, U.K.}
% \IEEEauthorblockC{}% <-this % stops an unwanted space
% \thanks{Manuscript received July 08, 2025; revised - -, 2025. 
% Corresponding author: Ch M. Awais (email: chmuhammad.awais@phd.unipi.it).}
}

% The paper headers
\markboth{Arxiv Preprint, 2025}%
{Awais \MakeLowercase{\textit{et al.}}: A Classification-Aware Super-Resolution Framework for Ship Targets in SAR Imagery}

% \IEEEpubid{© 2025 The Authors. This work is licensed under a Creative Commons Attribution-NonCommercial-NoDerivatives 4.0 License.}
% Remember, if you use this you must call \IEEEpubidadjcol in the second
% column for its text to clear the IEEEpubid mark.

\maketitle

\begin{abstract}
High-resolution imagery plays a critical role in improving the performance of visual recognition tasks such as classification, detection, and segmentation. In many domains, including remote sensing and surveillance, low-resolution images can limit the accuracy of automated analysis. To address this, super-resolution (SR) techniques have been widely adopted to attempt to reconstruct high-resolution images from low-resolution inputs. Related traditional approaches focus solely on enhancing image quality based on pixel-level metrics, leaving the relationship between super-resolved image fidelity and downstream classification performance largely underexplored. This raises a key question: can integrating classification objectives directly into the super-resolution process further improve classification accuracy? In this paper, we try to respond to this question by investigating the relationship between super-resolution and classification through the deployment of a specialised algorithmic strategy. We propose a novel methodology that increases the resolution of synthetic aperture radar imagery by optimising loss functions that account for both image quality and classification performance. Our approach improves image quality, as measured by scientifically ascertained image quality indicators, while also enhancing classification accuracy.
\end{abstract}


\section{Introduction}\label{sec:intro}
Ship classification algorithms help identify vessels, a critical task given that approximately 80\% \cite{antony2022emerging} of the world's trade is carried out via maritime transport, making maritime traffic monitoring crucial. Ship classification can be performed using images captured by synthetic aperture radar (SAR) \cite{chaturvedi2012ship, lang2018ship}; however, SAR data processing and its effectiveness are frequently hindered by two significant challenges: data scarcity and inherent low resolution of publically available data collections \cite{7364165}. In addition to these factors, SAR images are affected by sidelobe interference, speckle, and angle-dependent scattering behaviour. Strong sidelobes generated by bright scatterers can mask weak targets and distort local structure \cite{xiang2024sidelobe}, while variations in azimuth can introduce nonlinear changes in scattering patterns \cite{xiang2025sar}.

\begin{figure}[h!]
    \centering
    \begin{subfigure}[c]{0.45\linewidth}
        \centering
        \begin{tikzpicture}
            \node[inner sep=0pt] (imgA) at (0,0)
                {\includegraphics[width=3.8cm]{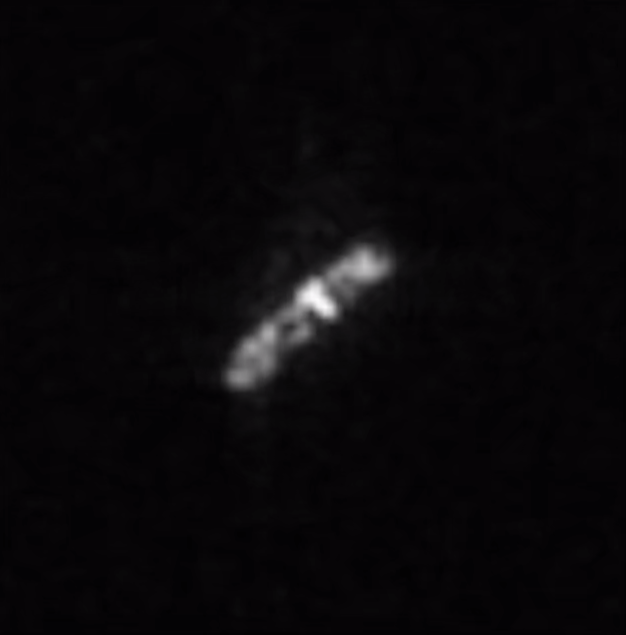}};
            \draw[green, thick, rounded corners] (-1.8,1.8) rectangle (1.8,-1.8);
            \node[align=center, text=green!80!black, font=\bfseries, above] at (0,1.9) {Focus on Whole Image};
        \end{tikzpicture}
        \caption{\small Traditional SR (Global Attention)}
    \end{subfigure}
    \hfill
    \begin{subfigure}[c]{0.45\linewidth}
        \centering
        \begin{tikzpicture}
            \node[inner sep=0pt] (imgB) at (0,0)
                {\includegraphics[width=3.8cm]{images/image_exmp.PNG}};
            \draw[green, thick, rotate around={35:(-0.25,-0.45)}] 
                (0.3,-0.1) ++(-0.8,0.3) rectangle ++(1.4,-0.6);
            % \draw[->, thick, green!80!black]
            %     (-1.3,1.95) .. controls (-0.8,1.0) and (-0.2,0.6) .. (-0.1,0.4);
            \node[align=center, text=green!80!black, font=\bfseries] at (-0.2,2.1) {Focus on ship target};
        \end{tikzpicture}
        \caption{\small Proposed (Localized Attention)}
    \end{subfigure}
    
    \caption{Comparison of attention strategies in super-resolution. Traditional methods apply global attention, while the proposed classification-aware approach uses class label supervision to guide attention toward task-relevant regions, enhancing super-resolution for downstream classification.}
    \label{fig:graph_abs}
\end{figure}

% Deep learning networks are effective due to their ability to learn from past examples, super resolution being a deep learning technique also uses the attribute to enhance the resolution of images \cite{anwar2020deep}.

Deep learning supervised networks are powerful algorithms that effectively learn from large amounts of annotated data. Such methods can also be applied to solve super resolution (SR) problems \cite{anwar2020deep}. %However, task-aware SR methods developed for optical images do not directly transfer to SAR, as SAR data contain speckle noise, sidelobes, and sea-clutter backgrounds that alter feature distributions. Preserving task-relevant scattering cues is therefore more challenging than in optical settings.
%can be enhanced using super-resolution (SR) techniques, which are effective due to their ability to learn from past examples \cite{anwar2020deep}. 
%%%% ORIGINAL SENTENCE
%Typically, pretrained SR models are trained on optical data, where they perform well and yield good results, then they are used for other applications. Researchers applied SR techniques to enhance SAR image quality, which is typically measured via metrics such as structural similarity index measure (SSIM) and peak signal-to-noise ratio (PSNR) \cite{yanshan2022ogsrn, 8634345, 6187672, zhang2023blind}. However, does this increase in image quality necessarily lead to better performance for applications like classification? The goal of enhancement is to improve SSIM and PSNR scores, but for classification purposes it is important to evaluate whether higher image quality translates into better performance for the said application.  
%%%%% Alternate version
SR models are typically pretrained on large-scale optical image datasets, where they have demonstrated considerable success. Following this pretraining, these models are often repurposed for specialized applications, such as enhancing SAR imagery. However, SR methods developed for optical images do not directly transfer to SAR, as SAR data contain speckle noise, sidelobes, and sea-cluttered backgrounds that alter feature distributions. In this context, the efficacy of SR techniques is typically quantified using pixel-based Image Quality (IQ) metrics, including the Structural Similarity Index Measure (SSIM) and Peak Signal-to-Noise Ratio (PSNR) \cite{yanshan2022ogsrn, 8634345, 6187672, zhang2023blind}. 

While these metrics may indicate an improvement in perceptual image quality, it is crucial to consider their relevance for subsequent analytical tasks. This raises a pivotal question: does an enhancement in SR-driven image quality, as measured by conventional metrics like SSIM and PSNR, necessarily translate to improved performance in downstream applications such as image classification? This disconnect is particularly relevant in SAR, where polarimetric studies show that preserving scattering entropy and backscatter contrast is critical for discrimination tasks \cite{xiang2025oil}. When super-resolution models are trained solely on pixel-level reconstruction losses (e.g., L1, L2), they optimize for pixel level accuracy but often produce over-smoothed results. In SAR ship classification, where targets are small and the scene is dominated by background clutter, this smoothing can inadvertently degrade the fine-grained scattering features crucial for accurate classification.

% Additionally, particularly in SAR-based ship classification, where images are largely dominated by background clutter and target vessels occupy only a small portion of the scene, the practical benefit of general image quality enhancements for improving target classification performance is not taken for granted.

To the best of our knowledge, no prior work has introduced a classification-aware super-resolution strategy specifically for SAR ship classification. While previous studies have optimized SR using pixel-level image quality metrics, our framework combines super-resolution and classification objectives to maintain scattering integrity and semantic consistency, addressing noise and imbalance challenges inherent to SAR ship classification. This approach encourages preservation of discriminative scattering characteristics that are critical for ship-class discrimination, producing super-resolved imagery that maintains both visual fidelity and classification accuracy (see Fig. \ref{fig:graph_abs}). We validate this hypothesis through a systematic experimental progression: applying off-the-shelf ImageNet SR models (SR-I), training SR models on SAR data with standard reconstruction loss (SR-PT), and finally incorporating classification awareness into the SR training process (SR-FT). To rigorously assess the effectiveness of the proposed methodology, we conducted experiments using three SR models and five classification networks. Evaluations on the widely used OpenSARShip dataset \cite{8124929}, which is known for its low-resolution imagery and severe class imbalance, revealed that image quality enhancements contribute positively to classification accuracy.

Our key contributions are summarised as follows:
\begin{itemize}
    \item We introduce two novel loss functions specifically designed to guide the training of SR models.
    \item We propose a unified framework that integrates classification feedback into the SR training process, enabling task-aware image enhancement.
    \item We demonstrate that the proposed approach simultaneously improves both image super-resolution quality and downstream classification performance.
\end{itemize}

Next sections are organized as follows: \cref{sec:background} provides a brief overview of the background and motivation for this research, \cref{sec:methodology} details the proposed methodology, \cref{sec:results} presents the results, \cref{sec:discussion} analyzes the findings, and we conclude our study in \cref{sec:conclusion}.

\section{Background}
\label{sec:background}
High-resolution images play a critical role in deep learning algorithms by preserving fine-grained spatial details that are essential for accurate feature extraction and robust model performance across visual recognition tasks. In SAR ship classification, where objects are small, high-resolution images provide richer visual details \cite{wang2018ship, 8094932, lu2018ship, dong2019fine, 9496207} that can be crucial for maritime applications, including traffic monitoring, maritime security, and environmental protection. However, SAR images often suffer from speckle noise and moderate resolution, leading to performance deficiencies in these applications. This underscores the need for techniques that enhance the resolution of SAR data while preserving critical details to improve ship classification accuracy.

Several examples exist in previous literature of how super-resolution techniques can be leveraged to achieve higher resolution in SAR imagery \cite{8094932, 9496207, karakucs2020solving}. Most researchers have focused on improving resolution by optimising image quality metrics such as PSNR \cite{lee2023efficient, jiang2024lightweight, 8899202, shen2020residual, 6187672} and SSIM \cite{lee2023efficient, jiang2024lightweight, 8899202, 8634345, 6187672}. Higher scores in these metrics provide evidence of the ability of SR models to improve image quality.

Typically, supervised SR machine learning algorithms are guided by the minimization of a loss function. This function estimates the difference between the super-resolved image generated by the model and  its corresponding ground-truth high-resolution counterpart;  the computed deviation is used to iteratively optimise the model's parameters. Also, it should be noted that the performance of an SR model, as assessed by PSNR and SSIM scores, depends on the choice of the loss function \cite{jo2020investigating, 10491161}. Pixel-wise losses, such as L1 and MSE, are commonly used to minimise reconstruction errors.

Despite the significance of SR in predicting high-resolution imagery and its applications in SAR data, its use in SAR ship classification remains scarcely explored. While PSNR and SSIM are widely used for image quality evaluation, they have not been incorporated as part of a loss function. The ultimate goal of generating high-resolution images is to enhance the performance of downstream tasks such as classification, detection, and segmentation. However, the majority of SR research remains focused on improving visual or perceptual image quality, often overlooking its practical impact on task-specific performance. In this study, we address this gap by integrating SR into the SAR ship classification pipeline and explicitly incorporating classification feedback into the SR training process. This enables the SR model to be optimised not only for visual quality but also for improving downstream classification accuracy.

\section{Methodology}
\label{sec:methodology}
\begin{figure*}
    \centering
    \includegraphics[width=1\linewidth]{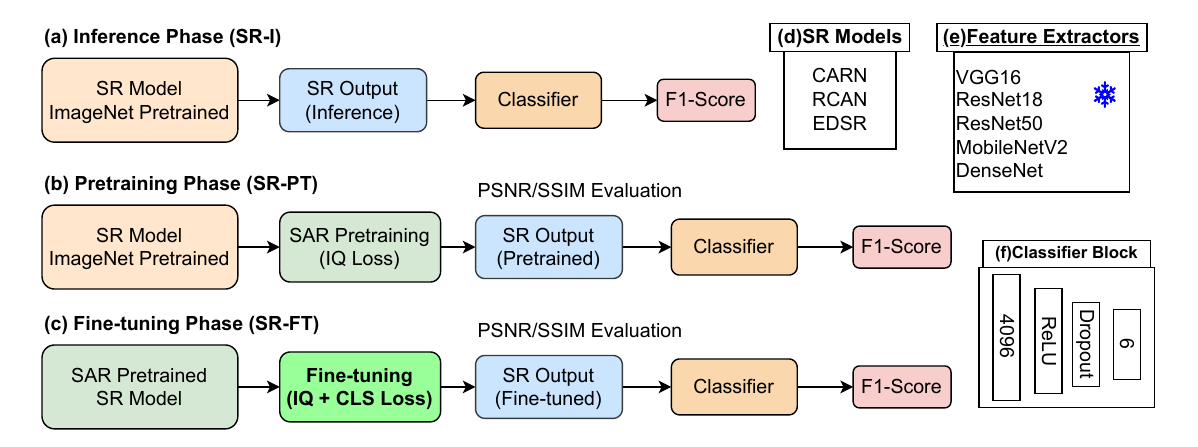}
    \caption{\textbf{Overview of the proposed classification-aware super-resolution pipeline} illustrating the \textbf{(a) SR-I}: inference using ImageNet pretrained SR models, \textbf{(b) SR-PT}: pretraining the SR models on SAR data using Image Quality (IQ) loss functions, and \textbf{(c) SR-FT}: finetuning on SAR data using IQ and classification focused (CLS) loss functions. Evaluation includes PSNR, SSIM, and F1-score. \textbf{(d) SR Models}: SR models used in the pipeline. \textbf{(e) Feature Extractors}: feature extractors used with frozen weights. \textbf{(f) Classifier Block}: classifier layers used for ship classification, where 6 is the number of OpenSARShip classes.}
    \label{fig:cvpr_methodology}
\end{figure*}

We divide our methodology into three stages, as illustrated in~\cref{fig:cvpr_methodology}. In the first stage (SR-I), we apply ImageNet pretrained SR models to infer high-resolution images and evaluate their performance using classification scores. The second stage (SR-PT) involves training the same SR models inferred in SR-I block on SAR data using loss functions driven by image quality metrics. In the third stage (SR-FT), we perform fine-tuning of the SR models pretrained in the SR-PT stage by incorporating a classification-guided loss function, enabling the model to optimise for both visual quality and classification performance. Each component of this methodology is described in detail below.

\subsection{Dataset and Models}
We utilized the publicly available OpenSARShip dataset \cite{8124929}, characterized by low-resolution synthetic aperture radar imagery. All classes within OpenSARShip were employed for training and evaluating the SR models; however, for the classification training and evaluation, we specifically selected six classes (Cargo, Tanker, Fishing, Dredging, Passenger, and Tug). Typically, prior studies have limited their analyses to three to five classes due to significant class imbalance issues. We expanded the analysis to six classes to robustly demonstrate the effectiveness of our proposed multi-stage methodology.

The dataset was preprocessed to create two subsets: one containing high-resolution (HR) original images and another consisting of low-resolution (LR) images generated by downsampling the original images by a factor of two. Three pretrained deep-learning-based SR models \cite{torchsr, ninasr} were selected to perform initial resolution enhancement: Enhanced Deep Super-Resolution (EDSR), Cascading Residual Network (CARN), and Residual Channel Attention Network (RCAN) (\cref{fig:cvpr_methodology}(d)). Subsequently, classification performance was evaluated using five widely recognized architectures (\cref{fig:cvpr_methodology}(d)): ResNet50, ResNet18 \cite{he2016deep}, VGG16 \cite{simonyan2014very}, MobileNetV2 \cite{sandler2018mobilenetv2}, and DenseNet121 \cite{huang2017densely}. The classification layers  (\cref{fig:cvpr_methodology}(e)) of these architectures were modified appropriately, while their feature extraction layers were fine-tuned without freezing to optimize both visual quality and classification accuracy.

\subsection{SR-I}
\begin{figure}[ht]
    \centering
    \includegraphics[width=1\linewidth]{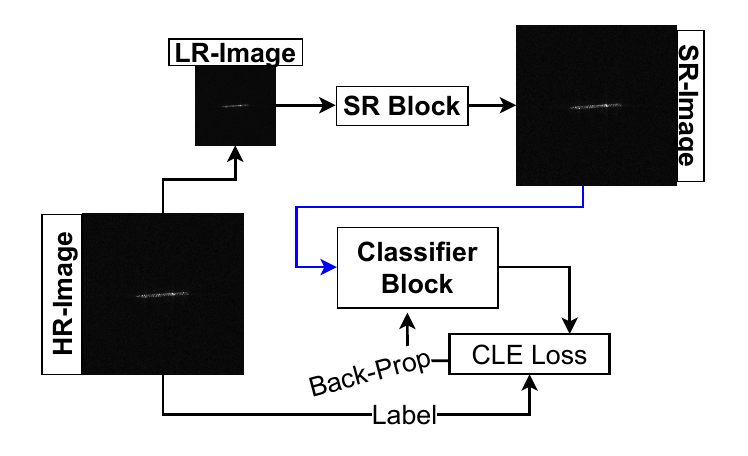}
    \caption{Overview of the SR-I stage}
    \label{fig:SR-I}
\end{figure}
To test the effectiveness of the SR models, in the inference module (\cref{fig:SR-I}), we utilized the SR-I models to generate SR images from LR images. Then, to evaluate the quality of images in terms of downstream tasks, we used the obtained SR images to train classification networks. The results are evaluated in terms of F1-scores.

\subsection{SR-PT}
\begin{figure}[h]
    \centering
    \includegraphics[width=1\linewidth]{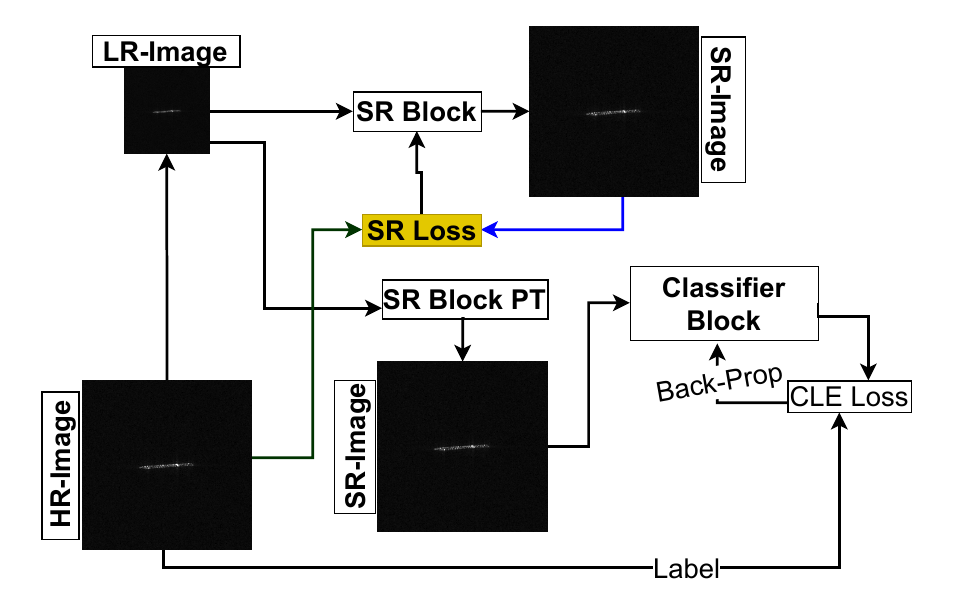}
    \caption{Overview of the SR-PT stage}
    \label{fig:sr_pt}
\end{figure}
In the SR-PT block (\cref{fig:sr_pt}), the SR models considered in previous block SR-I are trained on all classes of our dataset. As before, the SR images produced by these trained networks are then used to train the cascaded classification network. Three different loss functions, which are discussed in detail as follows, are used to train the SR models:
\subsubsection{L1-Loss}
The first adopted loss function is the $L1$ \cite{he2022revisiting}, which is defined in \cref{eq:l1Loss}:
\begin{equation}
\label{eq:l1Loss}
%L_1(hr, sr) = \frac{1}{N} \sum_{i=1}^{N} \left| sr^{(p)}i - hr_i \right|
%alt version (no reference to images in the left hand side in a consistent way with other equations)
L_1 = \frac{1}{N} \sum_{p=1}^{N} \left| sr^{(p)} - hr^{(p)} \right|
\end{equation}
where $hr^{(p)}$ is the $p$-th original image, $sr^{(p)}$ is the corresponding super resolved image, and $N$ is the total number of images.
\subsubsection{Combo-Loss}
As previously noted, PSNR and SSIM are commonly used to assess the quality of super-resolved images, but they are rarely leveraged to directly guide the training of super-resolution models. To address this, we propose the ``Combo-Loss'' function (\cref{eq:comboLoss}) that integrates both metrics to explicitly steer the training process toward generating perceptually and quantitatively improved images. The loss is defined as a weighted sum of a PSNR- and an SSIM-based terms:

\begin{equation}
L_{\text{combo}} = \alpha L_{\text{PSNR}} + \beta L_{\text{SSIM}}
\label{eq:comboLoss}
\end{equation}

% PSNR \cite{4775883} is defined as 

% \[
% PSNR = 10\log_{10}\!\left(\frac{M^2}{\text{MSE} + \epsilon}\right)
% \]

% where $M$ is the maximum possible value in the image, $\epsilon$ is a small constant (e.g., \(10^{-8}\)) to prevent division by zero and $\text{MSE}$ is defined as
% \[
%     \text{MSE(sr,hr)} = \frac{1}{N_p}\sum_{i=1}^{N_p} \left( sr_i - hr_i \right)^2, 
% \]

% with $N_p$ being the total number of pixels.
 
% According to this definition, $PSNR$ is upper bounded by 

% \[
% PSNR_{max} = 10 \log_{10}\!\left(\frac{M^2}{\epsilon}\right),
% \]

% Hence the PSNR term in the loss function is defined as

% \begin{equation}
%      L_{\text{PSNR}} = \frac{PSNR_{max} - PSNR}{PSNR_{max}}
% \end{equation}

PSNR is taken from \cite{4775883} and is used to define the loss function as follows:
\begin{equation}
     L_{\text{PSNR}} = \frac{PSNR_{max} - PSNR}{PSNR_{max}}
\end{equation}

$PSNR_{max}$ is the maximum $PSNR$ value and is given by 

\[
PSNR_{max} = 10 \log_{10}\!\left(\frac{M^2}{\epsilon}\right),
\]

where $M$ is the maximum possible intensity value in the image and $\epsilon$ is a small constant (\(10^{-8}\) was used in this work) to prevent division by zero.
% Particularly, the PSNR-based term is derived by converting the PSNR value into a loss:
% \begin{equation}
%      L_{\text{PSNR}} = norm(-10 \log_{10}\!\left(\frac{1}{\text{MSE} + \epsilon}\right))
% \end{equation}

% \begin{equation}
%     \text{MSE} = \frac{1}{N}\sum_{i=1}^{N} \left( sr_i - hr_i \right)^2,
% \end{equation}
% where \(norm\) is for normalizing psnr value, and \(\epsilon\) is a small constant (e.g., \(10^{-8}\)) to prevent division by zero.

Concerning the SSIM-based term, it is worth considering that the SSIM between any two images is a real number ranging from 0 to 1 (respectively indicating null to perfect similarity). Thus, given a set of N couples of super-resolved and high-resolution images, the SSIM \cite{4775883} term in the loss function (which penalises lower similarity) is defined as:
\begin{equation}
        L_{\text{SSIM}} = \frac{1}{N}\sum_{p=1}^{N} \Bigl(1 - \text{SSIM}(sr^{(p)},\, hr^{(p)})\Bigr)
\end{equation}

The resulting combination loss is formulated as a weighted average of the PSNR and SSIM losses, as shown in Equation~\ref{eq:comboLoss}. In the experiments conducted in this study, the weighting parameters were set to $\alpha = 0.5$ and $\beta = 0.5$, assigning equal importance to both components.
We determined the internal weights of the Combo loss through quantitative evaluation of PSNR, SSIM, and classification F1-scores. Equal weights of 0.5 and 0.5 provided the best balance between structural fidelity and signal preservation, and therefore this configuration was adopted in our experiments.

\subsubsection{Hybrid-Loss}
% original
% The hybrid loss function was introduced to assess whether combining traditional loss with image quality loss improves performance. It is designed to balance pixel-wise accuracy, global fidelity, and structural perception for SAR super-resolution. 
% alternative
To investigate whether integrating image quality metrics with a traditional pixel-level loss can enhance model performance, a hybrid loss function was developed. This function is formulated to strike a balance between pixel-wise accuracy, structural integrity, and perceptual quality. The proposed hybrid loss is a weighted sum of three components, respectively related to L1, SSIM and PSNR loss terms.

% The proposed hybrid loss is a weighted sum of three distinct components.
% L1 loss is assigned the highest weight (0.7) to preserve fine-grained structural integrity, SSIM is given a moderate weight (0.2) to enhance overall visual quality, and PSNR is assigned the lowest weight (0.1) to avoid overemphasis on pixel-wise accuracy, which may not always align with SAR-specific features. These weights were empirically chosen through extensive experimentation across  multiple datasets to achieve optimal performance. 
We determined the internal weights of the Hybrid loss by evaluating six different weight permutations of the $(0.7, 0.2, 0.1)$ configuration across L1, SSIM, and PSNR. The arrangement that assigned the highest weight to L1, a moderate weight to SSIM, and the lowest weight to PSNR was found to yield the most stable performance in both image quality and classification accuracy.

L1 loss is assigned the highest weight (0.7) to preserve fine-grained structural integrity, SSIM is given a moderate weight (0.2) to enhance overall visual quality, and PSNR is assigned the lowest weight (0.1) to avoid overemphasis on pixel-wise accuracy, which may not always align with SAR-specific features. The precise formulation of hybrid loss is presented in \cref{eq:hybridLoss}:

\begin{equation}
    \label{eq:hybridLoss}
    L_{\text{hybrid}} = 0.7 L_1 + 0.2 L_{\text{SSIM}} + 0.1 L_{\text{PSNR}}
\end{equation}

\subsection{SR-FT}

\begin{figure}[h!]
\centering
\adjustbox{scale=0.80,center}{ % Adjust scale here for single-column fit
\begin{tikzpicture}[
    node distance=1.6cm and 1.3cm,
    every node/.style={font=\small},
    box/.style={draw, thick, minimum width=2.3cm, minimum height=1cm, align=center, fill=blue!10},
    image/.style={draw, thick, minimum width=1.5cm, minimum height=1cm, align=center, fill=gray!10},
    loss/.style={draw, thick, ellipse, fill=red!10, minimum width=2.1cm, align=center}
]

% Nodes
% \node[image] (input) {LR Image};
\node[box] (srmodel) {SR-PT\\Model};
\node[image, right=of srmodel] (sroutput) {SR Image};

\node[box, below=1.3cm of sroutput] (classifier) {Classifier};
\node[loss, right=of sroutput] (srLoss) {SR Loss};
\node[loss, right=of classifier] (clsLoss) {Classification\\Loss};

% \node[draw, dashed, thick, fit=(srLoss) (clsLoss), inner sep=6pt, label=below:{\textbf{Joint Loss}}, fill=yellow!10] (jointLoss) {};
\node[draw, dashed, thick, fit=(srLoss) (clsLoss), inner sep=6pt, label=below:{\textbf{Joint Loss}}] (jointLoss) {};

% Arrows
% \draw[->, thick] (input) -- (srmodel);
\draw[->, thick] (srmodel) -- (sroutput);
\draw[->, thick] (sroutput) -- (srLoss);
\draw[->, thick] (sroutput) -- (classifier);
\draw[->, thick] (classifier) -- (clsLoss);

% Optimization arrows
\draw[<-{Latex[length=3mm]}, thick, dashed, red] (srLoss.north) -- ++(0,0.6) -| (srmodel.north);
\draw[<-{Latex[length=3mm]}, thick, dashed, red] (clsLoss.south) -- ++(0,-0.6) -| (srmodel.south);

\end{tikzpicture}
}
% \caption{Closer look of finetuning block}
\caption{Architecture of the fine-tuning (SR-FT) stage. The SR model, pre-trained for perceptual quality during SR-PT stage, is fine-tuned using a joint loss function that combines super-resolution loss and task-specific classification loss.
% This dual-objective optimization ensures that the enhanced images not only exhibit high visual fidelity but also contribute to improved downstream classification performance.
}
\label{fig:sr_ft}
\end{figure}

In the fine-tuning block (\cref{fig:sr_ft}) resolution is increased through transfer learning by leveraging SR models initially trained using image quality-focused losses in the SR-PT block. We fine-tune the SR models to further refine their ability to reconstruct high-resolution details. %During the SR training, we also calculate the classification loss of the SR images by comparing the classification result of both SR and HR images obtained by employing the five classification architectures mentioned in section III.A. %The classification layers of these five models are modified, while their feature extractor layers are used without freezing them.
A key aspect of this fine-tuning process is the integration of a task-specific objective. Alongside the SR refinement, we compute a classification loss by passing both the generated SR images and the ground-truth HR images through the five classification architectures mentioned in section III.A. The feature extraction layers of these classifiers remain trainable (i.e., unfrozen) to allow them to adapt to the specific image domain, while their final classification layers haven been modified for the specific task. Mainly (\cref{fig:cvpr_methodology}(f)), we have used a two-layer feedforward head comprising a linear transformation to 4096 units, followed by a ReLU activation and dropout regularization, and a final linear layer projecting to 6 output values, i.e. the chosen number of target classes. This parallel computation of classification loss allows the SR model's fine-tuning to be guided not only by image reconstruction fidelity but also by the ultimate goal of improving downstream classification performance.

\begin{algorithm}[htbp]
%\caption{Finetuning Block}
\caption{Pseudo-code describing the calculation of the loss in the SR-FT block. The resulting loss function incorporates super-resolution and classification penalty terms}
\label{alg:fine_tuning_block}
\begin{algorithmic}[1]
\State \textbf{Input:} $ hr, lr, \text{sr\_model}, \text{cls\_model}, \text{criteria} $
\State \textbf{Output:} $ merged\_loss $
\State Criteria $ \text{sr\_criterion}, \text{cls\_criterion} \gets \text{criteria} $
\State Compute SR image: $ sr \gets \text{sr\_model(}lr\text{)} $
% \State Initialize classification loss: $ cls\_losses \gets 0.0 $
% \For{$ cls\_model $ in $ \text{classifier\_models} $}
\State HR prediction: $ out\_hr \gets cls\_model(hr) $
\State SR prediction: $ out\_sr \gets cls\_model(sr) $
\State CLS loss: $ cl\_loss \gets \text{cls\_criterion(}out\_sr, out\_hr\text{)} $
\State SR loss: $ sr\_loss \gets \text{sr\_criterion(}sr, hr\text{)} $
\State $ merged\_loss \gets cls\_loss + sr\_loss $
\end{algorithmic}
\end{algorithm}

One specific novelty of this work concerns the definition of the loss function employed to train the models in this block (see Algorithm~\ref{alg:fine_tuning_block}). To force the SR models to take into consideration the peculiar information intrinsic in the categorical label of the training data, a novel factor is introduced in the loss, which exploits this information to constrain the classifier prediction. In particular, the SR images and HR images are fed into the five different classifiers, and the loss is computed by comparing the predicted labels of SR images with those of HR images:
% \[ L_{CLS} = \mathrm{mismatch\_error}(SR,HR)\]
(\cref{eq:mergedLoss}):
\begin{equation}
    \label{eq:mergedLoss}
    % L_{merged} = SR_{loss} + CLS_{loss}
    L_{merged} = L_{SR} + L_{CLS}
\end{equation}

% where $L_{CLS}$ is MSE Loss Function \cite{4775883}. 
% \[L_n = SR_{loss} + CLS_{loss}\] 

where $L_{SR}$  represents the loss from the SR training block, and $L_{CLS}$ is the mean-squared error (MSE) \cite{4775883} loss between the predicted labels.

Once more, following the fine-tuning process, the SR images produced by this SR-FT block are used to train and evaluate the suite of classification models. Performance is then quantified by calculating the final F1-scores for classification and the corresponding IQ metrics, ensuring consistency with the evaluation protocol of the previous stages.

\subsection{Training Settings}

For our experiment we use the OpenSARShip dataset, where LR images are generated by downscaling the original images to 32×32, while HR images have a resolution of 64×64. The learning rate for both model types (classification and SR) is set to 0.0001, and training is conducted for 10 epochs, with 64 batch size. Adam is used as the optimiser, and random seeds are set to ensure reproducibility. The implementation details, including access to the code, are available online \url{github.com/cm-awais/sar_classification_informed_sr}.

\begin{table*}[htbp]
\centering
\caption{Main evaluation of SR methods (EDSR, CARN, RCAN) across training stages (baseline, pretraining, fine-tuning) using PSNR (dB), SSIM, and F1-score. F1-scores are averaged across five classification models. \textbf{Bold} indicates the best per column (training phase); \cellcolor{green!40}green marks overall best performance for each metric. Averages per model are shown at the bottom, and best average is highlighted using \textbf{bold} font.}
\resizebox{0.95\textwidth}{!}{%
\begin{tabular}{ll|ccc|ccc|ccc}
\toprule
\textbf{SR Method} & \textbf{Loss} & \multicolumn{3}{c|}{\textbf{PSNR (↑)}} & \multicolumn{3}{c|}{\textbf{SSIM (↑)}} & \multicolumn{3}{c}{\textbf{F1-score (↑)}} \\
& & SR-I & SR-PT & SR-FT & SR-I & SR-PT & SR-FT & SR-I & SR-PT & SR-FT \\
\midrule

\rowcolor{gray!20}
  & Baseline LR & - & - & - & - & - & - & 61.664 & - & - \\
  & Baseline HR & - & - & - & - & - & - & 62.614 & - & - \\
\bottomrule
\multirow{4}{*}{EDSR}
  & Baseline & 42.47 & - & - & 0.97 & - & - & 48.718 & - & - \\
  & \textit{L1}     & - & 43.95 & 37.60 & - & 0.9762 & 0.9030 & - & 60.122 & 58.754 \\
  & \textit{Combo}  & - & 43.82 & 37.33 & - & 0.9708 & 0.8970 & - & 59.99 & 60.914 \\
  & \textit{Hybrid} & - & 40.98 & 35.91 & - & 0.9334 & 0.8290 & - & 60.902 & 59.558 \\
\rowcolor{gray!20}
  & \textbf{Average} & 42.47 & 42.92 & 36.95 & 0.97 & 0.9601 & 0.8763 & 48.718 & 60.338 & 59.742 \\
\midrule
\multirow{4}{*}{CARN}
  & Baseline & \textbf{42.71} & - & - & \cellcolor{green!40}\textbf{0.98} & - & - & 57.938 & - & - \\
  & \textit{L1}     & - & 42.87 & 37.34 & - & 0.9757 & 0.9070 & - & 60.324 & 60.844 \\
  & \textit{Combo}  & - & 43.23 & \textbf{38.14} & - & 0.9734 & \textbf{0.9260} & - & 59.73 & 60.47 \\
  & \textit{Hybrid} & - & 43.01 & 37.95 & - & 0.9696 & 0.9220 & - & 60.668 & 61.166 \\
\rowcolor{gray!20}
  & \textbf{Average} & 42.71 & 43.04 & 37.81 & \textbf{0.98} & 0.9729 & 0.9183 & 57.938 & 60.241 & 60.827 \\
\midrule
\multirow{4}{*}{RCAN}
  & Baseline & 42.66 & - & - & 0.97 & - & - & \textbf{58.826} & - & - \\
  & \textit{L1}     & - & 43.75 & 37.48 & - & 0.9760 & 0.9190 & - & 60.594 & 62.04 \\
  & \textit{Combo}  & - & \cellcolor{green!40}\textbf{44.31} & 37.75 & - & 0.9756 & 0.9050 & - & \textbf{61.18} & \cellcolor{green!40}\textbf{63.41} \\
  & \textit{Hybrid} & - & 43.42 & 37.75 & - & 0.9678 & 0.9100 & - & 60.022 & 63.15 \\
\rowcolor{gray!20}
  & \textbf{Average} & 42.66 & \textbf{43.83} & 37.66 & 0.97 & 0.9731 & 0.9113 & 58.826 & 60.599 & \textbf{62.867} \\
\bottomrule
\end{tabular}}
\label{tab:integrated_analysis}
\end{table*}

\section{Results}
\label{sec:results}
This section focuses on the impact of fine-tuning SR models using classification-aware loss functions and is structured into three main parts. First, we assess the performance of various SR algorithms using image quality metrics and downstream classification accuracy. Second, we compare different loss functions to evaluate their effectiveness in improving image fidelity and classification outcomes. Finally, we analyse the performance of multiple classification architectures to determine the most suitable models for integration with SR frameworks.

\subsection{SR Models}
% Throughout the first experimental analysis, the baseline represents the scores calculated without training the SR models on the SAR data (SR-I), whilst pretrained (SR-PT) and fine-tuned (SR-FT) refer to the scores calculated using the proposed methodology mentioned in \cref{fig:cvpr_methodology}.

% In terms of PSNR (\cref{tab:integrated_analysis}), the RCAN model optimised with the Combo-Loss in SR-PT block achieved the highest score. Overall, SR-PT models outperformed SR-FT and SR-I models in PSNR. For SSIM, the best result was obtained by Imagenet pretrained baseline (SR-I) with no training on SAR data, reaching a score of 0.98. However, when averaged across all methods, the SSIM scores were generally comparable, indicating consistent performance in structural similarity across different training strategies.

% To assess whether super resolved images lead to better classification performance, we evaluated the F1-score of the classification networks introduced in III.A, trained with the images generated by the several SR models proposed in sections III.b,III.c and III.d. The results indicate that RCAN with Combo-Loss, when fine-tuned in SR-FT block, achieved the highest F1-score. However, except for EDSR (possibly due to it being the only network without an adaptive mechanism), all other SR models showed improved performance during SR-FT compared to SR-PT, demonstrating the broad applicability of our proposed technique.

Throughout the first experimental analysis, the baseline represents the scores calculated without training the SR models on the SAR data (SR-I), whilst pretrained (SR-PT) and fine-tuned (SR-FT) refer to the scores calculated using the proposed methodology described in \cref{fig:cvpr_methodology}.

The evaluation of SR methods is conducted from two complementary perspectives: image quality assessment and classification performance. As summarised in \cref{tab:integrated_analysis}, SR-PT models generally achieved higher PSNR and SSIM values than their fine-tuned counterparts. RCAN obtained the highest PSNR across all loss functions in SR-PT (\cref{fig:cvpr_methodology}(b)), whereas EDSR consistently produced the lowest PSNR in both SR-PT and SR-FT (\cref{fig:cvpr_methodology}(c)). A similar trend was observed for SSIM, with RCAN outperforming the remaining models and EDSR ranking lowest across configurations. These results indicate that SAR-specific pretraining substantially enhances perceptual fidelity, even though fine-tuning prioritises classification relevance rather than pure image quality.

In terms of PSNR (\cref{tab:integrated_analysis}), the RCAN model optimised with Combo-Loss in the SR-PT block achieved the highest score. Overall, SR-PT models outperformed SR-FT and SR-I models in PSNR. For SSIM, the best score was obtained by the ImageNet-pretrained baseline (SR-I) with a value of 0.98. However, when averaged across all methods, SSIM values were generally similar, indicating consistent structural fidelity across different training strategies.

To assess whether super-resolved images lead to better classification performance, we evaluated the F1-score of the classification networks introduced in Section~III-A, using images generated by the SR models described in Sections~III-B, III-C and III-D. The results indicate that RCAN with Combo-Loss, when fine-tuned in the SR-FT block, achieved the highest F1-score. With the exception of EDSR (which may be limited by the absence of adaptive mechanisms), all other SR models showed improved performance during SR-FT compared to SR-PT, demonstrating the broad applicability of our proposed classification-aware strategy.

\subsection{Loss Functions}
The performance of image quality loss functions followed a similar trend to that of SR models in terms of PSNR and SSIM scores with SR-PT gaining better scores with respect to SR-FT. Consistently, SR-FT led to higher F1-score values across all loss functions (see \cref{tab:finetuning_effectiveness_loss}), and both Combo and Hybrid beat the L1 loss function for image quality.

% \begin{table}[htbp]
% \centering
% \resizebox{\columnwidth}{!}{%
% \begin{tabular}{lccc}
% \toprule
% \textbf{Loss Function} & \textbf{Pre-trained (PT)} & \textbf{Fine-tuned (FT)} & \textbf{Improvement} \\
% \midrule
% L1 & 60.35 & 60.56 & 0.21 \\
% \textbf{Combo} & 60.30 & \textbf{61.60} & \textbf{1.30} \\
% Hybrid & 60.53 & 61.20 & 0.67 \\
% \bottomrule
% \end{tabular}}
% \caption{Effectiveness of fine-tuning across loss functions, clearly highlighting Combo as the most effective loss function based on F1-score improvement.}
% \label{tab:finetuning_effectiveness_loss}
% \end{table}
\begin{table}[htbp]
\centering
\caption{The table presents a comparison of F1-scores (averaged across the employed classification models), demonstrating the impact of SR-FT on both classification performance and image quality, LR is low resolution data, HR is high resolution data, SRHR is the data which is super-resolved using SR-I models without any training on current dataset. The best scores for each loss functions are highlighted in bold.}
\renewcommand{\arraystretch}{1.1} % reduces row spacing slightly
\setlength{\tabcolsep}{4pt} % reduces horizontal padding
\resizebox{\columnwidth}{!}{% ensures table fits exactly the column width
\begin{tabular}{lcccccc}
\toprule
 & LR & HR & SRHR & L1 & Combo & Hybrid \\ 
\midrule
No Train   & 61.66 & 62.614 & 55.16 & -- & -- & -- \\ 
SR-PT & -- & -- & -- & 60.35 & 60.30 & 60.53 \\ 
SR-FT & -- & -- & -- & \textbf{60.56} & \textbf{61.60} & \textbf{61.29} \\ 
\bottomrule
\end{tabular}}
\label{tab:finetuning_effectiveness_loss}
\end{table}

While SR-PT optimises for perceptual fidelity, SR-FT shifts the optimisation toward discriminative feature recovery. This explains why the gain in F1-score is more pronounced for Combo and Hybrid losses: both incorporate structural and perceptual cues that translate more effectively into class-relevant features once classification gradients are introduced. In contrast, L1 focuses primarily on pixel-wise accuracy, which does not necessarily capture the scattering patterns or high-level semantics required for reliable ship discrimination. This behaviour is consistent with the trends observed across the SR models and highlights that loss functions incorporating perceptual structure, provide a stronger foundation for classification-aware fine-tuning.

\subsection{Classification Models}
To evaluate the impact of super resolution on classification, five different classification architectures were tested. VGG16 achieved the highest average F1-score (\cref{fig:cls_models}), while Resnet18 had the lowest. Although the performance difference among lower-performing models was minimal, VGG consistently outperformed others, making it the best-performing classification model for this task.

\begin{figure}[htbp]
    \centering
    \begin{tikzpicture}
        \begin{axis}[
            ybar,
            symbolic x coords={ResNet18,VGG16,MobileNetv2,ResNet50,DenseNet121},
            xtick=data,
            nodes near coords,
            ymin=55,ymax=65,
            ylabel={F1-score},
            xlabel={Models},
            bar width=0.7cm,
            width=0.99\columnwidth,
            height=6cm,
            enlarge x limits=0.2,
            xticklabel style={rotate=45,anchor=east}
            ]
            \addplot coordinates {(ResNet18,56.786) (VGG16,62.123) (MobileNetv2,58.032) (ResNet50,60.692) (DenseNet121,59.417)};
        \end{axis}
    \end{tikzpicture}
    \caption{Average F1-scores of different models clearly indicating \textbf{VGG16} as the best-performing model.}
    \label{fig:cls_models}
\end{figure}

% We also applied the three-stage super-resolution process to the MSTAR vehicle dataset \cite{keydel1996mstar}. The SR-I and SR-PT stages produced low F1-scores (10–17 \%), while SR-FT achieved the best results among the SR variants. This outcome reflects the large domain differences between ground and maritime SAR imagery and confirms that classification-aware fine-tuning provides the most stable adaptation. %Complete per-model results are reported in the Supplementary Material.

\begin{figure}[ht]
    \centering
    \includegraphics[width=1\linewidth]{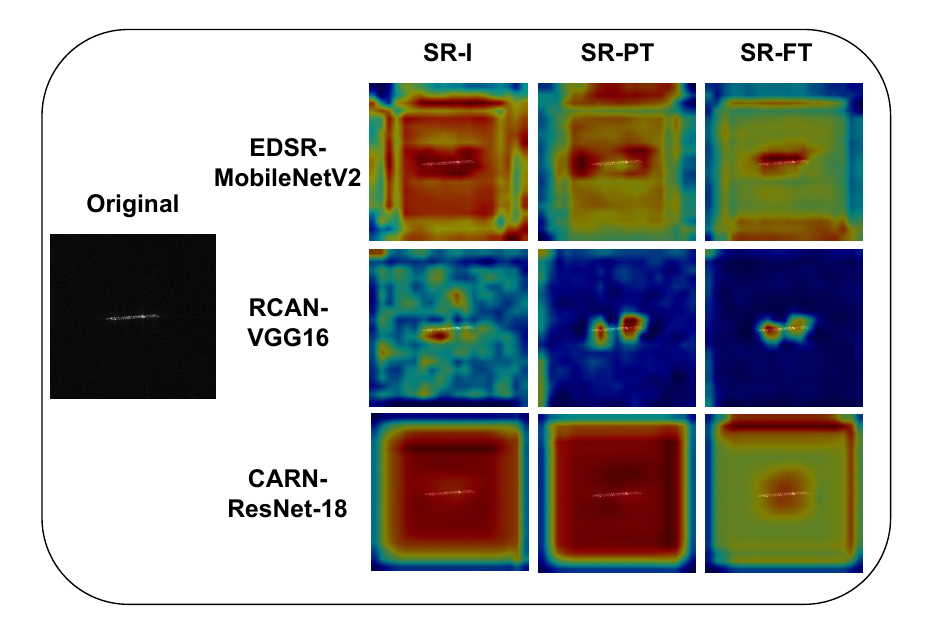}
    \caption{Grad-CAM visualizations illustrating classifier attention for SR-I, SR-PT, and SR-FT across different SR–classifier combinations. Warmer colors indicate stronger contributions to the classifier's decision. The SR-FT stage produces compact, target-centered activations focused on the ship, demonstrating improved task relevance.}
    \label{fig:gradcam}
\end{figure}

To further illustrate how the three SR stages affect the classifier’s decision-making process, we provide Grad-CAM visualizations in Fig.~\ref{fig:gradcam}. These show how SR-I and SR-PT tend to focus on background or noisy regions, whereas SR-FT yields compact, target-centered activations aligned with the ship structure.

% \subsection{Overall Effect}
% The overall performance results demonstrate that fine-tuning leads to improved PSNR, SSIM, and F1-scores (\cref{tab:training_effect}), confirming the effectiveness of our approach.

% \begin{table}[htbp]
% \centering
% \renewcommand{\arraystretch}{1.1} % reduces row spacing slightly
% \setlength{\tabcolsep}{4pt} % reduces horizontal padding
% \resizebox{\columnwidth}{!}{% ensures table fits exactly the column width
% \begin{tabular}{lcccccc}
% \toprule
%  & LR & HR & SRHR & L1 & Combo & Hybrid \\ 
% \midrule
% No Train   & 61.66 & 64.35 & 55.16 & -- & -- & -- \\ 
% Pretrained & -- & -- & -- & 60.35 & 60.30 & 60.53 \\ 
% Fine-tuned & -- & -- & -- & \textbf{60.56} & \textbf{61.60} & \textbf{61.20} \\ 
% \bottomrule
% \end{tabular}}
% \caption{The table presents F1-scores comparison, demonstrating the impact of fine-tuning on both classification performance and image quality, LR is low resolution data, HR is high resolution data, SRHR is the data which is resoloved using pretrained SR models without any training on current dataset. The best scores for each metric are highlighted in bold, and second best are underlined.}
% \label{tab:training_effect}
% \end{table}

\section{Discussion}
\label{sec:discussion}
% The discussion section provides valuable information on the selection of SR models for the classification of SAR ships, the identification of the most effective loss functions, and the determination of the classification models that perform best in such scenarios.

This section discusses the relationship between image quality, classification accuracy, and the behavior of different SR models under the proposed training pipeline. We first examine the effect of resolution scale on both image quality and classification behavior.

The 32×32 to 64×64 configuration represents a low-information regime that is typical in publicly available SAR ship datasets, where targets occupy a very small number of pixels and fine details are inherently limited. Although visual improvements at this scale are subtle, the classification-aware fine-tuning stage consistently enhanced the discriminative features most relevant to the classifier. Similar trends were observed when scaling from 64×64 to 128×128, demonstrating that the proposed approach remains effective across different resolution levels. This further highlights that the primary objective of the framework is to improve task-relevant feature representations rather than purely perceptual quality.

\subsection{SR Models}
Although one might expect classification outcomes to mirror image quality trends, our results show a clear divergence. The only exception was EDSR trained using L1 and Hybrid losses, where a marginal drop in F1-score was observed. All other models showed consistent improvements in classification performance after fine-tuning. For example, RCAN with Combo Loss improved from an F1-score of 61.18\% to 63.41\% (\cref{tab:integrated_analysis}). These findings confirm our earlier hypothesis \cite{11043629} that higher image quality, as measured by PSNR and SSIM, does not necessarily translate into better classification performance.
%These results suggest that higher image quality, as measured by PSNR and SSIM, does not always lead to better classification performance, hence proving our previous hypothesis \cite{11043629}.

A possible explanation of this behavior can refer to the nature of SAR imagery. SAR images often contain speckle and structural noise. SR methods can unintentionally enhance these noisy regions, especially if the model lacks mechanisms to distinguish signal from noise. EDSR is effective at preserving global image structure due to its deep residual architecture. However, it lacks adaptive mechanisms, such as attention modules or multi-scale feature fusion, which are found in RCAN and CARN. As a result, EDSR struggles to adapt to the semantic characteristics of SAR data during SR-FT (finetuning). This leads to suboptimal performance on the classification task.

In contrast, RCAN and CARN include architectural components (such as residual cascades and channel attention) that may help them focus on class-relevant features (e.g. ship contours and high-frequency textures). They achieve this even if it comes at the cost of traditional image quality metrics. This suggests a trade-off, where preserving semantic features becomes more important than maximizing pixel-level accuracy. These findings highlight the need to evaluate SR models based not only on reconstruction metrics but also on their performance in downstream tasks. For SAR ship classification, where there is a lot of noise, fine-tuning is essential to optimize class-discriminative features, even if it results in lower PSNR or SSIM values.

\begin{table*}[htbp]
\centering
\caption{\textbf{Best Performing combinations}: Classification F1-scores for the best-performing SR–classifier configurations. LR reports performance on low resolution inputs; HR denotes performance on the original high resolution images. SR-PT and SR-FT show F1 scores after applying 2× super resolution to HR images using the pretrained and fine-tuned SR models, respectively. Improvement gives the absolute increase in F1-score achieved by fine-tuning (SR-FT - SR-PT). Boldface highlights the highest fine-tuned score for each classifier.}
\resizebox{0.8\textwidth}{!}{%
\begin{tabular}{lcccccc}
\toprule
\textbf{Model (Loss)} & \textbf{LR} & \textbf{HR} & \textbf{SR-PT} & \textbf{SR-FT} & \textbf{Improvement} \\
\midrule
VGG16 (CARN-Combo)       & 61.95 & 65.03 & 63.12 & \textbf{65.40} & +2.28 \\
MobileNetv2 (RCAN-Combo) & 58.92 & 63.73 & 60.84 & \textbf{62.40} & +1.56 \\
MobileNetv2 (RCAN-L1)    & 58.92 & 63.73 & 60.56 & \textbf{61.35} & +0.79 \\
ResNet50 (EDSR-Combo)    & 61.22 & 60.40 & 62.34 & \textbf{62.85} & +0.51 \\
\bottomrule
\end{tabular}}
\label{tab:finetuning_improvements}
\end{table*}

\subsection{Loss Functions}
Also, in the context of loss functions (\cref{tab:finetuning_effectiveness_loss}), it is critical to identify which configuration yields the best performance for SAR ship classification. Since the trends observed in image quality metrics align with those seen across different SR models, our analysis focuses on their influence on classification performance, specifically the F1-score.

To evaluate this, we used both SR-PT (\cref{fig:cvpr_methodology}(b)) and SR-FT (\cref{fig:cvpr_methodology}(c)) models to generate SR images, which were subsequently used to train classification networks. The resulting F1-scores were analyzed to assess the relative effectiveness of different SR loss functions. Eventually, it was observed that SR images coming from models trained with Combo and Hybrid loss functions (see section \ref{sec:methodology}-C) were consistently classified better than those coming from L1-based SR models. %Combo Loss with VGG16 and CARN emerged as the best-performing configuration (\cref{tab:finetuning_improvements} represents the results on HR images when they are super resolved to 2x).
Combo Loss with VGG16 and CARN emerged as the best-performing configuration (see \cref{tab:finetuning_improvements}, which reports classification on HR images after 2× super-resolution).

An important trend was observed following fine-tuning with classification loss (SR-FT): all models exhibited improved F1-scores compared to their pre-trained counterparts. This indicates that while all three loss functions performed comparably during the pretraining phase, Combo and Hybrid losses provided a significant advantage during fine-tuning, confirming their suitability for enhancing classification performance in SAR applications.

To demonstrate the effectiveness of the proposed loss functions, additional comparative experiments have been performed taking into consideration several different training datasets FUSAR \cite{hou2020fusar}, MSTAR \cite{keydel1996mstar}, and HRSID \cite{9127939}. For FUSAR and MSTAR, performance was averaged across both SR-PT and ST-FT phases. For HRSID, only the SR-PT stage results were considered, as this dataset is designed for object detection rather than classification. The models were evaluated on each dataset using standard image quality metrics, including PSNR and SSIM.

\subsection{Classification Models}

The classification model exhibited lower scores when trained solely during SR-PT with SR loss functions (\cref{tab:integrated_analysis}), as shown in previous results. We attribute this to OpenSARShip being the lowest-resolution dataset, leaving the model with minimal information for training. Additionally, lower resolution causes fewer distinguishable features in images. To address this, classification loss was introduced during fine-tuning, which effectively improved performance in both target quality and classification scores.

% Another part of our methodology was to pretrain and fine-tune the network with all publicly available SAR ship data (Fusar, HRSID, and MSTAR) to determine whether this could further enhance OpenSARShip performance. However, since the classification models exhibited underfitting on other datasets (with F1-scores below 20\%) whilst achieving significantly higher F1-scores ($>$57\%) on OpenSARShip, using only OpenSARShip data yielded better results. Therefore, we exclusively used the OpenSARShip dataset. Also, we could not use other datasets in our methodology due to the limitation of the GPU, as other datasets are already in better resolution (FUSAR and HRSID with 256x256 patch size, and MSTAR with 128x128 patch size).

% Another part of our methodology was to explore the pretraining and fine-tuning of the network using all publicly available SAR ship datasets (FUSAR, HRSID, and MSTAR) to investigate whether this could enhance OpenSARShip performance. However, due to GPU limitations, we were unable to include these higher-resolution datasets in our training pipeline (FUSAR and HRSID use 256×256 patches, and MSTAR uses 128×128). Additionally, when we evaluated our classification models on these datasets, the performance was poor (F1-scores below 20\%), indicating underfitting, whereas the models achieved significantly better results on OpenSARShip (F1-score $>$ 57\%). These factors led us to exclusively use the OpenSARShip dataset in our methodology.
\begin{figure*}[htbp]
    \centering
    \includegraphics[width=0.65\linewidth]{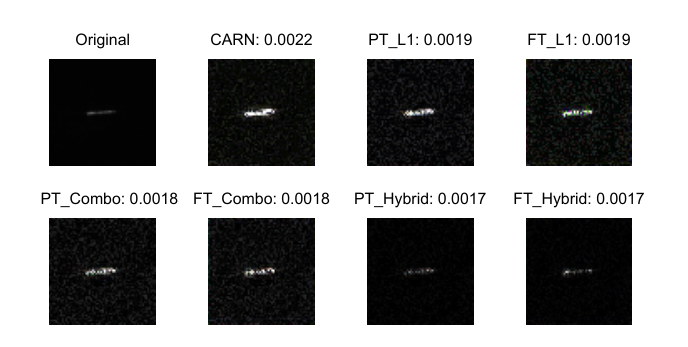}
    \caption{
    Error visualisation of super-resolved images using CARN. The images represent the pixel-wise difference between the original image and the SR outputs, with brighter regions indicating areas where the SR models failed to reconstruct details accurately. PT corresponds to the SR image generated by the model from the SR training block, while FT represents the SR image produced by the Fine-Tuned model. Each image includes a numerical error score, where higher values indicate greater reconstruction errors and poorer SR performance.}
    \label{fig:opensar_rcan_diff}
\end{figure*}
% To further analyse the effectiveness of different loss functions, we computed error maps (\cref{fig:opensar_rcan_diff}) by subtracting the original OpenSARShip image from the corresponding super-resolved images. This visualisation allows us to identify the regions where SR models struggle to reconstruct details accurately. The results indicate that the hybrid and Combo-Loss functions produced lower error rates compared to raw and L1 loss. Notably, errors were concentrated in the ship target area, suggesting that SR models fail to focus effectively on SAR targets. By incorporating classification loss into SR training, we mitigated this limitation, enabling the models to better capture SAR-specific features.
The divergence between image quality metrics and classification performance underscores a fundamental limitation in current evaluation methodologies for SAR super-resolution. Traditional metrics like PSNR and SSIM, while valuable for natural image assessment, may not adequately capture the preservation of discriminative features essential for SAR ship classification. Our error analysis (\cref{fig:opensar_rcan_diff}) supports this hypothesis, revealing that reconstruction errors are predominantly concentrated in ship target regions, precisely where accurate feature preservation is most critical for classification performance. This spatial distribution of errors suggests that conventional super-resolution approaches may struggle to maintain the subtle textural and structural characteristics that distinguish different ship classes in SAR imagery.

Fine-tuning enabled the SR models to better adapt to the underlying data distribution, resulting in consistent classification gains compared to pretraining alone. By integrating classification supervision into the SR training process, our framework allows the SR model to optimize for task-relevant features

\section{Conclusion}
\label{sec:conclusion}

% This paper presents the first comprehensive evaluation of SAR super-resolution for ship classification while integrating classification loss into SR training. The proposed methodology enhances SAR image quality by introducing image quality-focused loss functions and improves classification performance by incorporating classification constraints. Experimental results demonstrate the effectiveness of this approach, yielding improved image quality and classification accuracy. In future work, we aim to extend this novel methodology to other computer vision applications, including object detection and segmentation tasks.

This paper presented a classification-aware super-resolution framework for SAR ship classification, combining image-quality losses with a classification-guided objective. The proposed methodology improves both perceptual image quality and downstream classification accuracy. Experiments conducted using three SR models and five classifiers demonstrated that incorporating classification information into the SR training process leads to consistent performance gains.

The main contributions of this work are as follows:
\begin{itemize}
    \item Two loss functions were introduced to guide SR models toward improved perceptual quality while preserving task-relevant structure.
    \item A comparative analysis identified CARN as the most effective SR model and VGG16 as the strongest classifier for SAR ship classification.
    \item It was shown that aligning SR optimization with classification metrics yields higher classification accuracy than traditional pixel-focused training.
\end{itemize}

Future work will explore extending this classification-aware SR strategy to additional SAR domains and tasks, such as detection and segmentation, by adapting the framework to task-specific architectures.

% \backmatter

% \bmhead{Supplementary information}

% If your article has accompanying supplementary file/s please state so here. 

% Authors reporting data from electrophoretic gels and blots should supply the full unprocessed scans for key as part of their Supplementary information. This may be requested by the editorial team/s if it is missing.

% Please refer to Journal-level guidance for any specific requirements.

% \bmhead{Acknowledgements}

% Acknowledgements are not compulsory. Where included they should be brief. Grant or contribution numbers may be acknowledged.

% Please refer to Journal-level guidance for any specific requirements.

\section*{Acknowledgments}
This work was supported by National Recovery and Resilience Plan (NRRP), Mission 4 Component 2 Investment 1.4 - Call for tender No. 3138 of 16 December 2021, rectified by Decree n.3175 of 18 December 2021  of Italian Ministry of University and Research funded by the European Union – NextGenerationEU. Award Number: Project code CN\_00000033, Concession Decree No. 1034  of 17 June 2022 adopted by the Italian Ministry of University and Research,  CUP D33C22000960007, Project title ``National Biodiversity Future Center - NBFC''.

% {\appendix[Proof of the Zonklar Equations]
% Use $\backslash${\tt{appendix}} if you have a single appendix:
% Do not use $\backslash${\tt{section}} anymore after $\backslash${\tt{appendix}}, only $\backslash${\tt{section*}}.
% If you have multiple appendixes use $\backslash${\tt{appendices}} then use $\backslash${\tt{section}} to start each appendix.
% You must declare a $\backslash${\tt{section}} before using any $\backslash${\tt{subsection}} or using $\backslash${\tt{label}} ($\backslash${\tt{appendices}} by itself
%  starts a section numbered zero.)}

%{\appendices
%\section*{Proof of the First Zonklar Equation}
%Appendix one text goes here.
% You can choose not to have a title for an appendix if you want by leaving the argument blank
%\section*{Proof of the Second Zonklar Equation}
%Appendix two text goes here.}

\bibliography{bare_jrnl_new_sample4.bib}
\bibliographystyle{IEEEtran} %IEEEtran-plainnat

\newpage

\section{Biography Section}
\vspace{-10pt}
\begin{IEEEbiography}[{\includegraphics[width=1in,height=1.15in,clip,keepaspectratio]{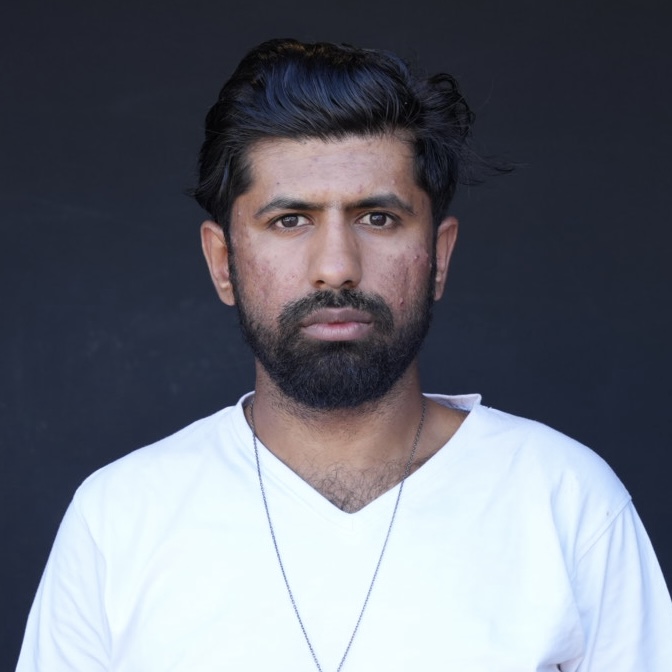}}]{Ch Muhammad Awais}
is a Ph.D. student at the University of Pisa, specializing in SAR ship classification. Hailing from Pakistan, he has been engaged in deep learning research for the past six years and has worked on SAR ship classification for the last two years.
\end{IEEEbiography}
\vspace{-40pt}
\begin{IEEEbiography}[{\includegraphics[width=1in,height=1.15in,clip,keepaspectratio]{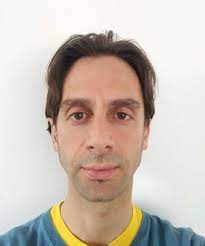}}]{Marco Reggiannini} received his M.Sc. in Physics and a Ph.D. in Automation Engineering from the University of Pisa. Currently, he is a researcher at the Signals and Imaging Lab at ISTI-CNR, with more than ten years of experience devoted to multisensor imagery analysis.
\end{IEEEbiography}
\vspace{-40pt}
\begin{IEEEbiography}[{\includegraphics[width=1in,height=1.15in,clip,keepaspectratio]{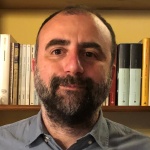}}]{ Davide Moroni} received his M.Sc. in Mathematics (Hons.) from the University of Pisa and a Ph.D. in Mathematics from the University of Rome La Sapienza. He is a Senior Researcher and Head of the Signals and Images Lab at ISTI-CNR, specialising in image processing and computer vision.
\end{IEEEbiography}
\vspace{-40pt}
\begin{IEEEbiography}[{\includegraphics[width=1in,height=1.25in,clip,keepaspectratio]{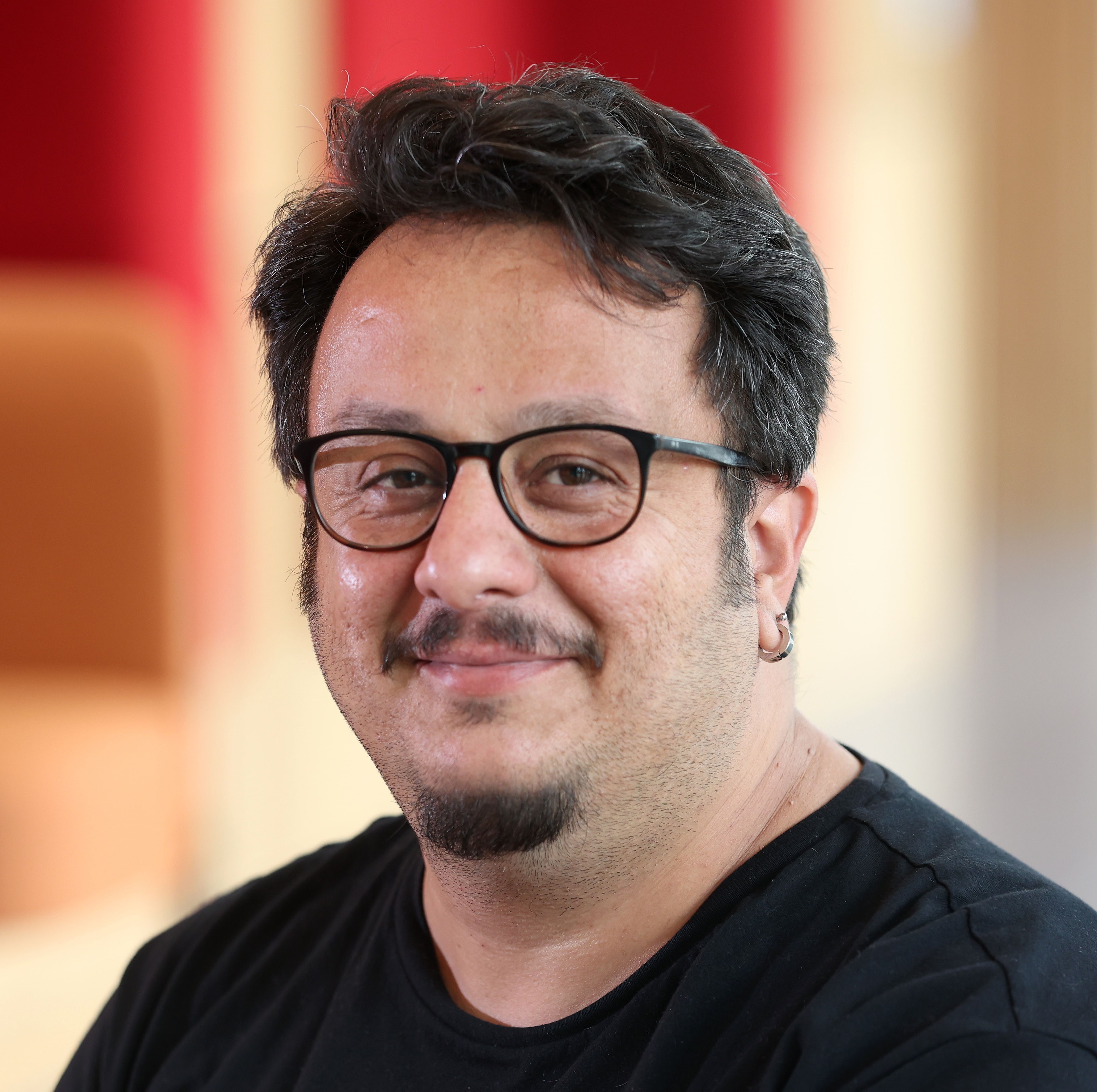}}]{Oktay Karakus}
 (Member, IEEE) received his B.Sc. (Hons.) degree in Electronics Engineering from Istanbul Kültür University, Turkey, in 2009, and his M.Sc. and Ph.D. degrees in Electronics and Communication Engineering from İzmir Institute of Technology (IZTECH), Turkey, in 2012 and 2018, respectively. From 2009 to 2018, he held research assistant positions at several universities in Turkey. In 2017, he was a Visiting Scholar at the Institute of Information Science and Technologies (ISTI-CNR), Pisa, Italy. From 2018 to 2021, Dr. Karakus was a Research Associate in image processing at the Visual Information Laboratory, University of Bristol, UK. He is currently a Lecturer at the School of Computer Science and Informatics, Cardiff University, where he also serves as Deputy Director of the Cardiff Data Science Academy and leads the Remote Sensing Image and Data Analysis (ReSIDA) research group. His research interests include remote sensing image analysis, environmental data science for marine and coastal sciences, ecology and wildlife, machine learning and AI, and football analytics. He is on the editorial board of multiple reputable journals and has published in high-impact venues in remote sensing, environmental monitoring, and artificial intelligence.
\end{IEEEbiography}

\vfill

\end{document}